\documentclass[a4paper,fleqn]{cas-dc}

\usepackage[numbers]{natbib}
\usepackage{booktabs}

\def\tsc#1{\csdef{#1}{\textsc{\lowercase{#1}}\xspace}}
\tsc{WGM}
\tsc{QE}
\tsc{EP}
\tsc{PMS}
\tsc{BEC}
\tsc{DE}

\begin{document}
\let\WriteBookmarks\relax
\def\floatpagepagefraction{1}
\def\textpagefraction{.001}
\shorttitle{ISA Transactions}
\shortauthors{Feiyan Min et~al.}

\title [mode = title]{$\mathscr{L}_1$ Adaptive Resonance Ratio Control for Series Elastic Actuator with Guaranteed Transient Performance}
\tnotemark[1,2]

\tnotetext[1]{This work was supported in part by the National Natural Science Foundation
of China under Grant 62172188}


\author[1]{Feiyan Min}
\cormark[1]

\credit{Conceptualization of this study, Methodology, Software, Writing}

\address[1]{Electronic engineering department, Jinan University, 510632, Guangzhou, China}

\author[1]{Gao Wang}[style=chinese]
\credit{Experiment}

\author[2]{Xueqin Chen}[%
   ]
\cormark[2]

\credit{Data curation, Supervision}

\address[2]{School of Astronautics, Harbin Institute of Technology, 150001, Harbin, China}



\cortext[cor1]{Corresponding author: Feiyan Min. E-mail address: feiyan min@jnu.edu.cn}
\cortext[cor2]{Co-corresponding author: Xueqin Chen. E-mail address: cxqhit@163.com}


\begin{abstract}
%
%
To eliminate the static error, overshoot, and vibration of the series elastic actuator (SEA) position control, the resonance ratio control (RRC)  algorithm is improved based on $\mathscr{L}_1$ adaptive control(L1AC)method. Based on the analysis of the factors affecting the control performance of SEA, the algorithm schema is proposed, the stability is proved, and the main control parameters are analyzed. The algorithm schema is further improved with gravity compensation, and the predicted error and reference error is reduced to guarantee transient performance.  Finally, the effectiveness of the algorithm is validated by simulation and platform experiments. The simulation and experiment results show that the algorithm has good adaptability, can improve transient control performance, and can handle effectively the static error,  overshoot, and vibration. In addition, when a link-side collision occurs, the algorithm automatically reduces the link speed and limits the motor current, thus protecting the humans and SEA itself, due to the low pass filter characterization of L1AC to disturbance.

\end{abstract}




\maketitle

\section{Introduction}

The series elastic actuator (SEA) was first proposed by Parrt in 1995 \cite{PrattWilliamson1995}, which can reduce the contact stiffness between the manipulator and the environment or human, to ensure the safety and adaptability of interactive operation for manipulators.
However, the introduction of series elastic actuators may increase the complexity of the robot dynamics model, deteriorate the transient performance, and lead to mechanical resonance, overshoot, and static error on the link side.

Series elastic actuators provide valuable properties compared to rigid joints \cite{PaineSentis2014}. First, the mechanical output impedance of the flexible joint is reduced, helping to establish a stable contact between the load and the motor output shaft. Second, the elastic element can absorb the energy impact generated by the accidental collision, reducing the peak moment of the collision and thus the impact on the mechanical structure. Third, elastic components can store mechanical kinetic energy, allowing higher peak power output. Fourthly, the elastic components can be used as a torque sensor due to the better linearity between the torsion angle and the output torque.

However, the position and force control of SEA is more complicated \cite{Calanca2016}. First of all, compared with the rigid structure, the dynamic model of SEA is more complex; the order of the model increases from 2 to 4, and the difficulty of stability and accuracy increase significantly. Furthermore, the elastic components are easy to cause vibration, which makes the link side still have residual jitter when it reaches the target position. Third, passive flexibility dramatically reduces the system's control bandwidth and limits the tracking ability.

Motion control of SEA is regarded as an essential issue in the next generation of industrial robots. Consequently, since the 1990s, the research of its control problem has become the focus of attention, and various control methods have been proposed.

Spong\cite{Spong1987}\cite{Spong1990}et al. proposed a simplified linear model of a flexible joint robot, and various control methods based on this model were researched. Tomei\cite{Tomei1991} successively proposed the PD control method and gravity compensation control strategy and proved the closed-loop stability condition. Although the PD control algorithm can guarantee convergence, there is no better demonstration of steady-state error and vibration.

Kouhei Ohnishi \cite{YukiOhnishi1993} \cite{SariyildizOhnishi2014} et al. proposed the Disturbance Observer (DOb) and Resonance Ratio Control (RRC) algorithm, which had a great attention. Among them, the disturbance observer is very effective in compensating for the friction of the motor side and has become a common algorithm for servo motors and industrial robot. The Resonance Ratio Control includes position control mode and force control mode \cite{ThaoYokokura2018}\cite{Mitsantisuk2013}\cite{Sariyildiz2021}. The idea is that through the state feedback of the motor side and the link side, the closed-loop poles are configured so that there are no conjugate poles, and the system response is non-oscillating. However, the algorithm assumes that the model and the external interference is precisely known, of which the latter is often difficult to satisfy.

In addition, a variety of control algorithms are proposed. The most common is the so-called linear model based control methods, including the Extended state observer (ESO) \cite{TalolePhadke2008} and feedback linearization \cite{Luca1998}, etc. Sun Lei \cite{Li2019}\cite{Sun2018} et al. proposed a controller with a non-linear structure to achieve effective suppression of overshooting and vibration, which was verified on a multi-link manipulator. And Daniel et al. propose the fractional-order control method for flexible joint robot\cite{Daniel2019}.

The ideal SEA position control should have adequate quick response speed, no static error, and good transient performance, that is, no overshoot and oscillation. On the other hand, when a collision occurs with the environment or humans, it is necessary to slow down and limit the motor current in time to protect humans, environment and the SEA itself.

For the affection of model uncertainty and environmental disturbance, the adaptive control method is worth considering. The commonly used model reference adaptive control (MRAC) \cite{Mohammad2021} methods need to establish a unified structure of parameter adaptive law and control law. Due to the high order of the SEA model, it is difficult to construct the complicated Lyapunov function.

Recently, the $\mathscr{L}_1$ adaptive method  \cite{Hovakimyan2009}\cite{Xargay2010}\cite{Lee2017}\cite{CaoHovakimyan2010} was proposed to achieve the separate design of parameter adaptive law and control law, thereby reducing the design difficulty and can be applied to high order system. The $\mathscr{L}_1$ adaptive method has been successfully applied to the control of aircraft,  AUV, and other industrial system, it has gained much attention\cite{Bing},\cite{Zhu2018},\cite{Yang2020}\cite{Zuo2015}.

In this paper, the $\mathscr{L}_1$ adaptive control (L1AC) method is used for the position control of SEA.
 For the SEA control, the main interference is located in the link side, while the drive is located in the motor motor side, this is an unmatch system \cite{Xargay2010}.

For this reason, we adopt the piecewise discrete adaptive law, which is based on linear state space model, and the nonlinear gravity is ignored in the initial design.
And then , we improve the structure of the standard L1AC with an online nonlinear gravity compensation algorithm, and the stability and tracking performance of the control algorithm are analyzed.

%
%
%

\section{Problem description}

A typical series elastic actuator that contains a motor, a torsional spring and additional link side inertia, and can be characterized with Spong's model \cite{Spong1987}.
\begin{equation}
\begin{array}{l}
\begin{cases}
  J_a \ddot{q}  + G = K_f(\theta - q) - \tau_{dis}\\
  J_m\ddot{\theta}  + \tau_{fm}(\theta,\dot{\theta}) + K_f(\theta -q) = \tau_m\\
\end{cases}
\end{array}
\end{equation}
where $J_a, G$ and $q$ are the inertia, nominal gravity and position of the link side, respectively, $J_m$, $\theta$ and $\tau_{fm}$ is the inertia, position and friction of motor side, respectively. $K_f$ is the spring coefficient. $\tau_{dis}$ is the disturbance force caused by the model uncertainty and the external environment.


The Resonance Ratio Control (RRC) proposed by Ohnishi \cite{YukiOhnishi1993} can eliminate the natural vibration of SEA well and has received wide attention, the control law is given by:
\begin{equation}
\begin{cases}
  u = K_p(\theta_d - \theta) - K_v\dot{\theta} + K_r(G-K_f(\theta -q)) \\
  \tau_m = J_m u + \hat{\tau}_{Dob}
\end{cases}
\end{equation}
where $\theta_d = q_d + G/K_f$ is the target position of motor side with gravity compensation, and $q_d$ is the target position of link side.

\begin{figure}[!t]
\centering
\includegraphics[width=3.55in]{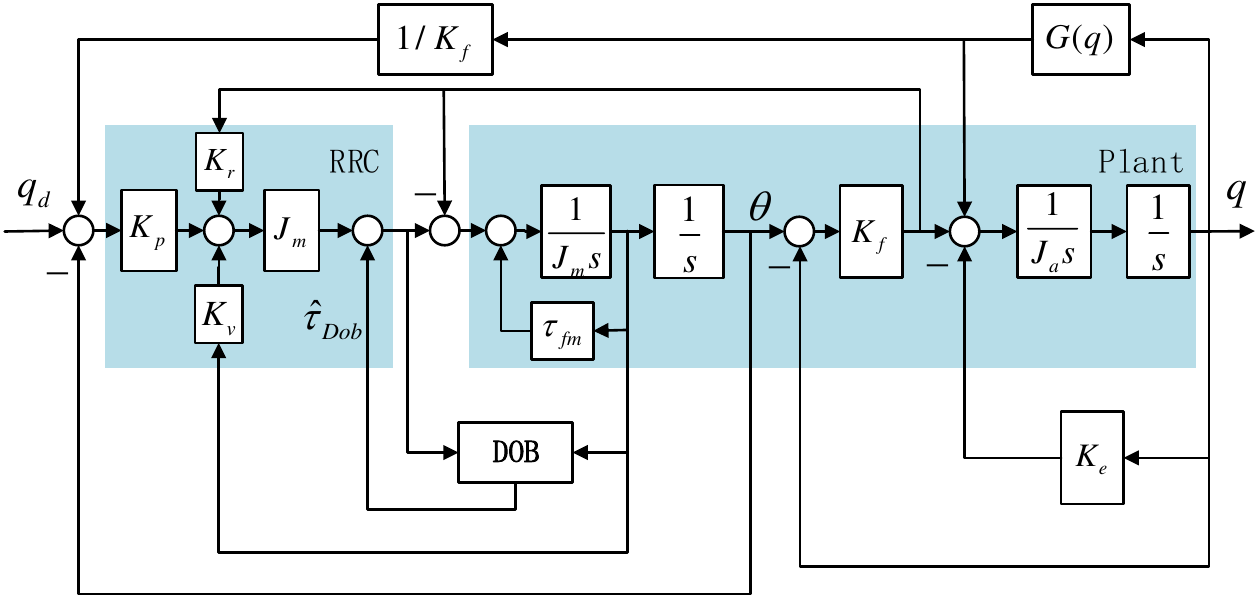}
\caption{The structure of standard resonance ratio control.}
\label{fig_1}
\end{figure}
The values of the other control parameters are
$ K_p = K_f/J_a$,
$ K_r = 4/J_a$,
$ K_v = 4\omega$ and $\omega = \sqrt{K_f/J_a}$.
And $\hat{\tau}_{Dob}$ is the output of the disturbance observer (DOb), which is used to compensate for motor side disturbance, including the mechanical friction and elastic force.

\begin{equation}
  \hat{\tau}_{Dob} = \frac{g_{ob}}{s + g_{ob}}(\tau_{fm}(s) + K_f(\theta(s)-q(s)))
\end{equation}
where $\tau_{fm}(s)$ is the Laplace Transform of $\tau_{fm}(\theta,\dot{\theta)}$. And $g_{ob}$ is the  bandwidth of the disturbance observer, which is much higher than that of the close-loop system.

The disturbance observer can compensate for the motor side interference well. Ignoring the estimation errors $\hat{\tau}_{Dob} - \tau_{fm} - K_f(\theta - q)$ of DOb, the response equation of the closed-loop system becomes

\begin{equation}
  \begin{cases}
    \ddot{q} &= K_f(\theta-q)/J_a-\tau_{dis}/J_a - G/J_a\\
    \ddot{\theta} & = K_p(\theta_d-\theta) - K_rK_f(\theta-q)  - K_v\dot{\theta} + K_rG\\
  \end{cases}
\end{equation}

If the link side disturbance $\tau_{dis}$ is further ignored and the gravity is fully compensated, the closed-loop response transfer function is
\begin{equation}
  q(s) = \frac{\omega^4}{(s + \omega)^4}q_d(s)
\end{equation}

In essence, the RRC implements online compensation for nonlinear terms $G$ and $\tau_{fm}(\theta,\dot{\theta})$ and assigns all closed-loop poles to $\omega$, and thus the overshoot and vibration of closed-loop response are eliminated.

However, the link side disturbance $\tau_{dis}$ will greatly damage the response performance.

In practice, the dominant disturbance can be described as $\tau_{dis} = \Delta G + K_e(q-q_0)$. The first term is the disturbance force caused by the uncertainty of the link side load.  The latter is the external force caused when the link is contact with the environment, $K_e$ is the environment stiffness and $q_0$ is the position when the link is contact with environment. And it follows that

\begin{equation}
   q(s) = \frac{\omega_a^4 q_d(s) -\Delta(s) + \rho_0(s)}{Q(s)}
\end{equation}
where $\Delta = -(s^2 + 4\omega s + 5\omega^2)\Delta G/J_a$, $\rho_0 = -(s^2 + 4\omega s + 5\omega^2)K_e q_0/J_a$, and close loop characteristic equation $Q(s)=s^4 + 4\omega s^3 + (6\omega^2+\lambda) s^2 + (4\omega^3 + 4\omega\lambda) s + (\omega^4+5\lambda\omega^2)$, $\lambda = K_e/J_a$.

The parameter root locus with respect to $\lambda$ is shown in the figure 2. Conjugate poles appear when $\lambda > 0$, so the overshoot and vibration increases with the increase of $K_e$.
Meanwhile, the mass uncertainty $\Delta G$ will introduce the steady-state error of the closed-loop response.

\begin{figure}[!t]
\centering
\includegraphics[width=3.0in]{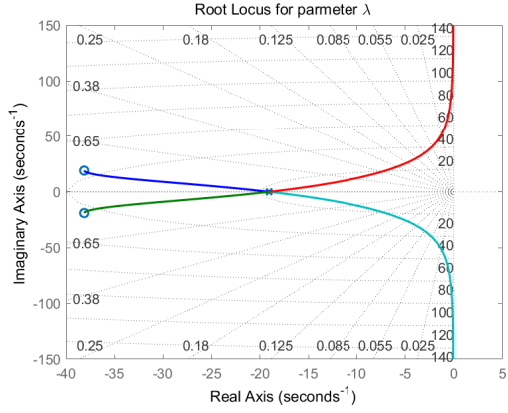}
\caption{Parametric root locus for different contact stiffness}
\label{fig_1}
\end{figure}

\section{$\mathscr{L}_1$ Adaptive motion control}
As discussed above, the main interference is located in the link side, while the drive is the  motor side, this is an unmatched system. For this structure, we adopt the piecewise discrete adaptive law in $\mathscr{L}_1$ controller design , which is based on linear state space model.

However, the nonlinear term gravity cannot be involved in the linear model, and is ignored as disturbance in the initial design. And then we improve the structure of controller with an online nonlinear gravity compensation algorithm, and the stability and tracking performance of the control algorithm are analyzed.

\subsection{$\mathscr{L}_1$ Adaptive resonance ratio control law}
The starting  $\mathscr{L}_1$ adaptive motion control system is given  by the following equation:
\begin{equation}
    \begin{cases}
      \dot{x}(t) & = A x(t) + B_m(u(t) + \sigma_1(t)) + B_{um} \sigma_2(t) \\
      x(0) & = x_0 \\
      y(t) & = c x(t)
    \end{cases}
\end{equation}
where $x(t)\in R^n$ is the state vector of the system. $A$ is constant system matrix, $B_m\in R^{n\times m}$ is the control channel matrix, $u(t)$ is the  control signal, $\sigma_1$ is the disturbance matched with $B_m$. $\sigma_2$ is the unmatched disturbance with channel matrix $B_{um}\in R^{n\times(n-m)}$, and $rank([B_m, B_{um}]) = n $.

\begin{figure*}
\centering
\includegraphics[width=6.0in]{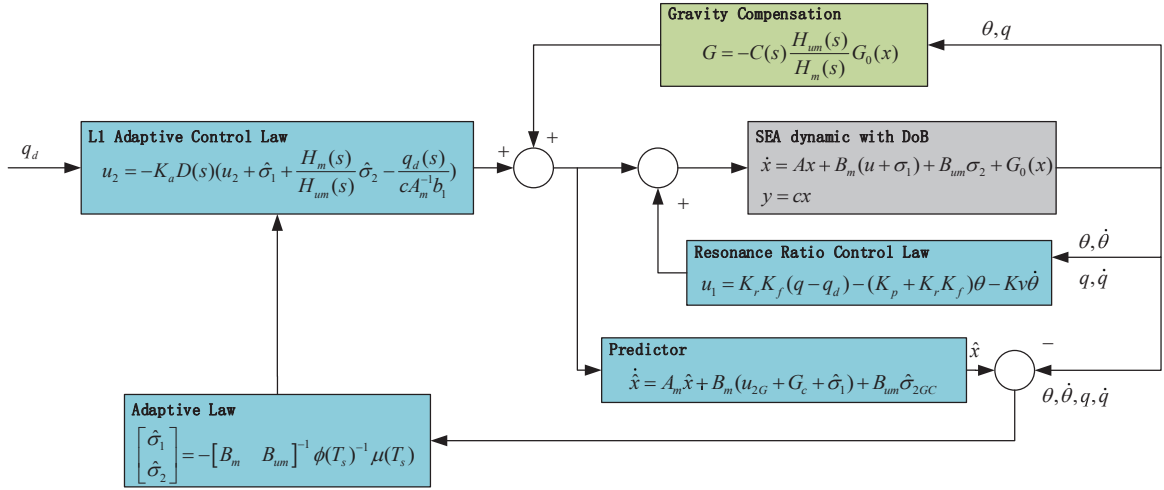}
\caption{$\mathscr{L}_1$ Adaptive Resonance Ratio Control for SEA}
\label{fig_1}
\end{figure*}

The control structure is given as follows:
\begin{equation}
  u(t) = u_1(t) + u_2(t)
\end{equation}
where $u_1(t)= -Kx(t)$ is the state feedback control law designed for the nominal system to realize poles assignment, and the  value of $K$ is required to make  $A_m = A - B_m K$ for Hurwitz.

Obviously, the RRC mentioned above is transformed into the state feedback control law $u_1(t)$. The state variable is chosen as $x = [q,\dot{q},\theta,\dot{\theta}]'$, and the control law in (2) is transformed as
\begin{equation}
 K = [-K_rK_f, 0, K_p+K_rK_f, K_v]
\end{equation}
where the nonlinear term $G$ is ignored in linear state feedback controller design. That means the gravity is considered as unknown disturbance in total and will be compensated in the adaptive control law $u_2(t)$. And the state equation is
\begin{equation}
  \begin{cases}
    \dot{x} = A_m x + B_m(u_2 + \sigma_1)+ B_{um} \sigma_2 \\
    A_m = \left[\begin{array}{cccc}
           0 & 1 & 0 & 0 \\
           -\frac{K_f}{J_a} & 0 & \frac{K_f}{J_a} & 0 \\
           0 & 0 & 0 & 1 \\
           K_rK_f & 0 & -K_rK_f-K_p & -K_v
         \end{array}\right] \\
     B_m = \left[\begin{array}{c}
                        0 \\
                        0 \\
                        0 \\
                        1
                      \end{array}\right]
                     \qquad B_{um} = \left[\begin{array}{ccc}
           1 & 0 & 0  \\
           0 & 1 & 0  \\
           0 & 0 & 1  \\
          0 & 0 & 0
         \end{array}\right]          \\
  \end{cases}
\end{equation}
 The adaptive control law $u_2(t)$  compensates the disturbance of  $\sigma_1, \sigma_2$ and tracks the target link side position $q_d$.

 The disturbance above is assumed to be bounded and verifies following condition:

Assumption 1:  $\sigma_1$ a scalar, there exit $L_1, B_1$ such that
\begin{equation}
  ||\sigma_1||_{\infty} \le L_1 ||x||_{\infty} + B_1
\end{equation}

Assumption 2: $\sigma_2 = [\sigma_{21},\sigma_{22},\sigma_{23}]'$ is a  three dimension column vector, and there exit $L_2, B_2$ such that
\begin{equation}
  ||\sigma_2||_{\infty} \le L_2 ||x||_{\infty} + B_2
  \end{equation}

\textbf{Remark 1.}
Among the four disturbances $\sigma_1, \sigma_{21},\sigma_{22},\sigma_{23}$, the motor side disturbance $\sigma_1$ is mainly the mechanical friction $\tau_{fm}$, and is compensated by the DOb. In contrast, the external force on the link side $\sigma_{22} = \tau_{dis}$ is the dominant disturbance for the SEA control.

To synthesize the adaptive law, it is necessary to predict the state and estimate the value of disturbance.
The state predictor equation is:
\begin{equation}
  \dot{\hat{x}} = A_m \hat{x} + B_m(u_2(t) + \hat{\sigma}_1) + B_{um}\hat{\sigma}_2
\end{equation}
where $\hat{x}, \hat{\sigma}_1, \hat{\sigma}_2$ is the predicting value of $x, \sigma_1, \sigma_2$, respectively.

 The adaption law estimates the matched disturbance $\sigma_1$ and unmatched disturbance $\sigma_2$, and the computational equation is as:
 \begin{equation}
   \begin{cases}
     \tilde{x}(iT_s) = \hat{x}(iT_s) - x(iT_s) \\
     \mu(T_s) = e^{A_m T_s} \tilde{x}(i T_s) \\
     \phi(T_s) = A_m^{-1}(e^{A_m T_s} - I_n)\\
     \left[\begin{array}{c}
             \hat{\sigma_1}((i+1)T_s) \\
             \hat{\sigma_2}((i+1)T_s)
           \end{array}\right] = -
                      \left[ B_m B_{um}
           \right]^{-1}\phi(T_s)^{-1}\mu(T_s)
   \end{cases}
 \end{equation}
where $\tilde{x}$ is the predict error, and $T_s$ is adaption rate.

 The adaptive control law $u_2(t)$ is updated with follow algorithm:
 \begin{equation}
   u_2  = -\frac{K_aD(s)}{(I_m +  K_aD(s)) } (\hat{\sigma}_1 + H_m^{-1}(s)H_{um}(s)\hat{\sigma}_2 - K_gq_d(s)) \\
 \end{equation}
 where the matrics $H_m(s)$, $H_{um}(s)$ and $K_g$ are defined as

\begin{equation}
  \begin{array}{l}
    H_m(s) = c(sI_n - A_m)^{-1}B_m \\
    H_{um}(s) = c(sI_n - A_m)^{-1}B_{um}\\
    K_g  = -(cA_m^{-1}B_m)^{-1}
  \end{array}
\end{equation}

The tuning parameters include the  low-pass filter $D(s)$ control gain $K_a$, they have to be chosen such that:
\begin{equation}
  C(s) =  K_aD(s)(I_m +  K_aD(s))^{-1}
\end{equation}
is stable and has DC gain $C(0) = I_m$; furthermore $C(s)H_m^{-1}$ has to be a strictly proper transfer function.

\subsection{Stability and Transient Performance}

Reference system is constructed for the proof of stability and transient performance of $\mathscr{L}_1$ adaptive control system. The reference system assumes perfect identification of all uncertainties, and cancel them within the bandwidth of the low pass filter $C(s)$. It should be noted that this auxiliary reference system is not implementable, and not involved in the implementation of the  $\mathscr{L}_1$ adaptive control system.
\begin{equation}
\begin{cases}
  \dot{x}_r = A_m x_r + B_m (u_{2r} + \sigma_{1r}) + B_{um}\sigma_{2r} \\
  u_{2r}(s) = -C(s)(\sigma_{1r} + H_{mum}(s)\sigma_{2r} - K_g q_{dr})
\end{cases}
\end{equation}
where  $u_{2r}(s)$ is the Laplace transform of the adaptive control law $u_2$, $H_{mum}(s) = H_m^{-1}(s)H_{um}(s)$, $x_r = [q_r,\dot{q}_r,\theta_r,\dot{\theta}_r]$ is the state variable of reference system. $\sigma_{1r}$ and $\sigma_{2r} = [\sigma_{21r},\sigma_{22r},\sigma_{23r}]'$ is the identified disturbance in the reference system.


The close loop transfer function of reference system is
\begin{equation}
  x_r(s) = G_1(s)\sigma_{1r} + G_2(s)\sigma_{2r} -G_d(s)K_g q_{rd}
\end{equation}
where
$G_1(s) = (sI_n - A_m)^{-1}B_m(1-C(s))$, $G_2(s) = (sI_n - A_m)^{-1}B_{um}(1-C(s))$ ,$G_d = (sI_n - A_m)^{-1}B_mC(s) $

For the proof of stability and transient performance bound of reference system, the choice of $K_a$ and $D(s)$ also needs to ensure that, there exits bounded positive $\rho_r$ such that the following condition holds:
\begin{equation}
  ||G_1||_{L1}l_0 + ||G_2||_{L1} < \frac{\rho_r + ||G_d||_{L1}||q_{dr}||_\infty}{L_2 \rho_r + B_0}
\end{equation}
where $l_0 = L_1/L_2$, $B_0 = max(B_1/l_0, B_2)$

\textbf{Lemma 1.} For the close loop reference system (18), subject to the condition(20), then
\begin{equation}
  ||x_r||_{\infty} \le \rho_r
\end{equation}

\textbf{Lemma 2.} For the prediction equation (13), (14), if there exits a time $\tau$ such that truncation function
 $ ||x_\tau||_\infty $ is bounded.
then
\begin{equation}
  ||\tilde{x}_{\tau}(t)||_\infty \le \Gamma
\end{equation}
where $\Gamma = (\alpha_1(T_s) + \alpha_2(T_s))\gamma_0 + \alpha_3(T_s)\gamma_1 +  \alpha_4(T_s)\gamma_2$

\textbf{Theorem 1.} Given the close loop system via (7)$\sim$(10), and the reference system in (18), if the condition is (20) holds, then
\begin{equation}
  \begin{array}{c}
     ||x_r - x||_\infty < \Gamma_r
  \end{array}
\end{equation}
where
$\Gamma_r = \frac{ ||G_d(s)H_m^{-1}(s)||_{L1} }{1-||G_1(s)||_{L1}L_1 -||G_2(s)||_{L1}L_2}\Gamma + \epsilon$,  and $\epsilon$ is an arbitrarily small positive constant.

\textbf{Remark 3.} The bound of reference error $\Gamma_r$  is also affected by the uncertainties and external disturbances $\sigma_1$ and $\sigma_2$, and can  be reduced
with small enough sampling time $T_s$.

\section{Controller Design}

This section designs a SEA controller based on the L1 adaptive control algorithm proposed above and compares it with the standard RRC control algorithm by simulation. Figure 4 shows the SEA test platform, and table 1 shows the relevant parameters. In order to test the adaptability of the algorithm to load changes, we add variable load to nominal link inertia. On the other hand, a collision test platform with variable stiffness is built to test the response characteristics of the position control algorithm under environmental collision.

\begin{table}
  \centering
  \caption{SEA system parameters.}
  \begin{tabular}{lll}
\toprule
 System parameters$\qquad$&  Actual value$\quad$& Unit    \\
 \midrule 
Motor side inertia&  $J_m=0.294$& $kg m^2$  \\
Link side inertia&  $J_a=0.345$& $kg m^2$ \\
\quad(With nominal load)& \quad & \\
 Spring stiffness& $K_f=125.478$ & $ N m/rad $ \\
  Load torque($q = 90^o$)& $G_0=8.856$ & $ N m  $ \\
  \quad (With nominal load)&  &  \\
  Nominal load & $m_0=1.5$ & $kg$ \\
Test load & $m = 0.5, 1.5, 2.0$ & $kg$\\
Motor torque coefficient & $K_t =0.094$ & $ N m/$\textperthousand  \\
Viscous coefficient & $f_m = 4.082$ & $N m /(rad/s)$ \\
Natural frequency& $\omega=19.068$ & $rad/s$ \\
Parameter of DoB& $g_{ob}=500$ & $rad/s$ \\
\bottomrule
\end{tabular}
\end{table}

The controller (8) includes state feedback control law $u_1$ and adaptive control law $u_2$. The $u_1$ adopts the control parameter of the standard RRC, as shown in (2) and (9). The $u_2$ form is as in (15), and the designed parameters include the adaptive gain $K_a$ and the low-pass filter $D(s)$ , which ensure that (17) is stable and $C(s)H_m^{-1}(s)$ is strictly proper. Since the SEA system is a fourth-order system, the candidate $D(s)$ are as follows
\begin{equation}
  D(s) = \frac{1}{s(T s+1)^3}
\end{equation}
where $T$ is the time constant of the filter. Since the sampling rate of the control system is $T_s = 0.001s$, $T$ should be greater than this value. Therefore, $T = 0.005$, 0.01, and 0.02s is selected for the test.

\begin{figure}[!t]
\centering
\includegraphics[width=3.55in]{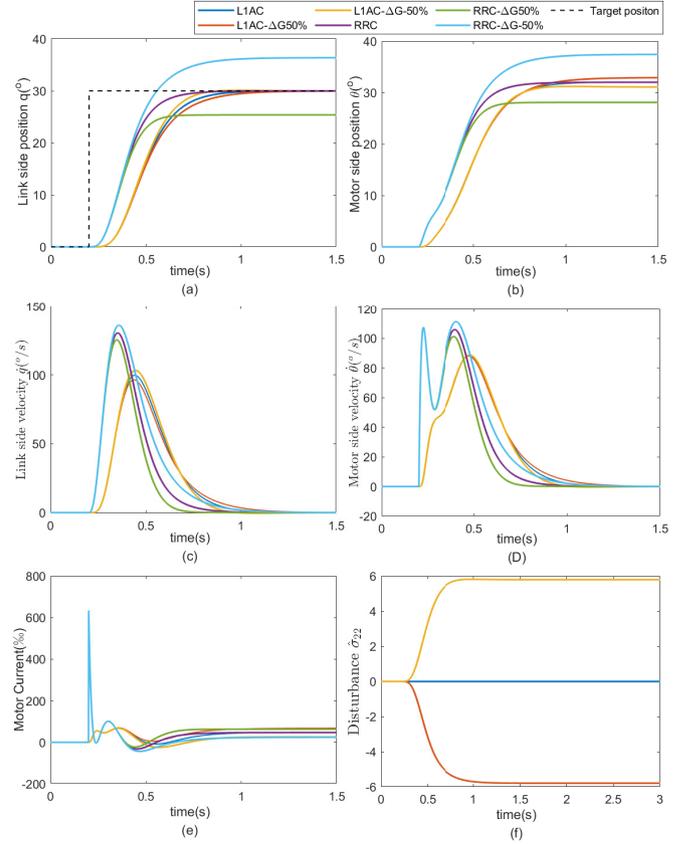}
\caption{Simulation results under load variation. The nominal mass is 1.5kg, and simulation experiments with loads of 0.75kg, 1.5kg, and 2.25kg are carried out using L1AC and RRC algorithms, respectively. The L1AC algorithm has no static error, but RRC has a sizeable static error. The response speed of L1AC is slow by 0.1s due to the influence of LPF $C(s)$, but there is no spike current in this algorithm, as in (e). $\hat{\sigma}_{22}$  in (f) is the estimate of the disturbance torque on the link side, which, multiplied by related coefficients, is the actual estimate of the external disturbance.
(The unit of motor current is one thousandth of the nominal current)}
\label{fig_1}
\end{figure}

\section{Conclusion}
This paper proposes a SEA control algorithm based on L1AC with a gravity compensation algorithm to improve the transient control performance. The main disturbance is located in the link side, while the drive is the  motor side, this is an unmatch system, and the piecewise discrete adaptive law is used, which is based on linear state space model. So the nonlinear term gravity is ignored as disturbance in the initial design. And then we improve the structure of controller with an online nonlinear gravity compensation algorithm, and the stability and tracking performance of the control algorithm are analyzed.

\appendix


\printcredits


\bibliographystyle{unsrt}

\bibliography{cas-refs}


%
%
%
%

\end{document}